\documentclass{article}

 \usepackage[final]{ARLET_2025}

\usepackage[utf8]{inputenc} % allow utf-8 input
\usepackage[T1]{fontenc}    % use 8-bit T1 fonts
\usepackage{hyperref}       % hyperlinks
\usepackage{url}            % simple URL typesetting
\usepackage{booktabs}       % professional-quality tables
\usepackage{amsfonts}       % blackboard math symbols
\usepackage{nicefrac}       % compact symbols for 1/2, etc.
\usepackage{microtype}      % microtypography
\usepackage{xcolor}         % colors
\usepackage{algorithm}
\usepackage{algpseudocode} % gives you \For, \If, \State, etc.
\usepackage{amsmath}

\usepackage{graphicx}  
\usepackage{listings}
\usepackage{xcolor}
\usepackage{listings}
\usepackage{makecell} % put this in preamble

\title{The Good, The Bad, and The Hybrid: A Reward Structure Showdown in Reasoning Models Training}

\author{%
  Subramanyam Sahoo\thanks{https://github.com/SubramanyamSahoo/Discrete-vs-Continuous-Rewards} \\
  Berkeley AI Safety Initiative (BASIS)\\
  UC Berkeley \\
  \texttt{sahoo2vec@gmail.com} \\
}

\begin{document}

\maketitle

\begin{abstract}
Reward design is central to reinforcement learning from human feedback (RLHF) and alignment research. In this work, we propose a unified framework to study \emph{hard}, \emph{continuous}, and \emph{hybrid} reward structures for fine-tuning large language models (LLMs) on mathematical reasoning tasks. Using Qwen3-4B with LoRA fine-tuning on the GSM8K dataset, we formalize and empirically evaluate reward formulations that incorporate correctness, perplexity, reasoning quality, and consistency. We introduce an adaptive hybrid reward scheduler that transitions between discrete and continuous signals, balancing exploration and stability. Our results show that hybrid reward structures improve convergence speed and training stability over purely hard or continuous approaches, offering insights for alignment via adaptive reward modeling.
\end{abstract}


\section{Introduction}
Recent advances in large language models have enabled impressive chain-of-thought (CoT) reasoning capabilities, where models generate explicit reasoning steps to solve complex tasks. For example, prompting a 540B model with CoT exemplars achieves state-of-the-art accuracy on GSM8K, a benchmark of grade-school math word problems. However, aligning LLMs to consistently produce correct reasoning remains challenging as these models often cannot reliably self-evaluate or refine their reasoning without external guidance. Reinforcement learning from human (or automated) feedback (RLHF) \cite{ouyang2022traininglanguagemodelsfollow} is a common approach to align LLM outputs to desired behaviors. In the RLHF pipeline, a pretrained LM is fine-tuned with an RL algorithm (e.g. PPO) using a scalar reward from a reward model (or other metric). In practice, specifying an appropriate reward signal for open-ended reasoning tasks is nontrivial. Simple scalar rewards (e.g. exact match accuracy) can be very sparse, whereas dense rewards (e.g. likelihood or intermediate metrics) may not align perfectly with the final goal. Moreover, reward models trained on preference data can generalize poorly out-of-distribution, especially on reasoning tasks. To address this, we explore adaptive reward strategies that combine multiple reward components and dynamically schedule their influence during training. Our key contributions are:

\textbf{Reward design}: We propose three reward formulations: (1) a hard binary reward (correct/incorrect + format bonus), (2) a continuous multi-component reward combining correctness, language model likelihood, reasoning quality, and consistency, and (3) hybrid combinations that transition between these. We provide detailed mathematical definitions for each (see Sec. 3).

\textbf{Hybrid scheduler}: We formalize a schedule for mixing hard vs continuous rewards. In the continuous-to-hard schedule, the training starts with full weight on the continuous reward and linearly shifts to the hard reward over a fixed number of steps; the hard-to-continuous schedule does the opposite. We give precise formulas for the time-dependent weights (Sec. 3.3).

\textbf{Evaluation on math reasoning}: We implement these reward schemes in a LoRA-fine-tuning framework \cite{hu2021loralowrankadaptationlarge}using Qwen3-4B \cite{yang2025qwen3technicalreport} and evaluate on GSM8K. We compare final accuracies and training dynamics. Notably, the hard reward achieved the best accuracy (40\%) and convergence, while the continuous reward lagged (28\%) despite richer signals. Hybrids performed moderately (33\%).

\textbf{Analysis and insights}: We analyze training stability and the contributions of each reward component, and discuss why continuous/hybrid schemes may underperform with naive weighting. We relate this to known alignment challenges: LLMs struggle to evaluate their own reasoning, and naive reward components (e.g. perplexity) may not capture true correctness. We also highlight how multi-objective reward models (e.g. multi-exemplar or consistency-based) could be integrated in our framework.

Our work emphasizes the importance of reward modeling in RL fine-tuning of LLMs, especially for open-ended tasks like math reasoning. We show how to systematically construct, formalize, and compare different reward structures and schedules. The framework is designed to be extensible (e.g. supporting new rewards or domains), and all training loop pseudocode, logging, and metrics are detailed for reproducibility.

\section{Related Work}
RLHF and alignment: Reinforcement learning has become a cornerstone for aligning LMs to human preferences, as popularized by ChatGPT and Claude. In typical RLHF, a reward model is trained to score outputs, and PPO is used to optimize the policy (LM) to maximize expected reward. Reward models themselves are often opaque and brittle: they may have only 60–70\% accuracy even on in-distribution tasks, and can fail on out-of-distribution examples. These issues are exacerbated for reasoning tasks, where ground-truth supervision is hard to obtain (e.g. verifying each reasoning step is laborious). Recent work (RewardBench
) \cite{lambert2024rewardbenchevaluatingrewardmodels} has begun to systematically evaluate RM generalization, finding shortcomings in reasoning and instruction-following. In short, RLHF is powerful but reward design remains a key challenge in aligning LMs.

Math reasoning and chain-of-thought: Datasets like GSM8K (8.5K grade-school math problems) and MATH (contest problems) are common benchmarks for evaluating LLM reasoning. Chain-of-thought prompting, which elicits intermediate reasoning steps, dramatically improves accuracy on GSM8K and similar tasks. For example, Wei et al. (2022) \cite{wei2023chainofthoughtpromptingelicitsreasoning} report that a 540B model with CoT prompts achieved state-of-the-art performance on GSM8K. Our method similarly encourages the model to produce an explicit reasoning segment (using XML tags), so that rewards can penalize incoherent or implausible chains. However, whereas CoT prompting relies on handcrafted prompts or exemplars, our approach uses RL to fine-tune the model with learned rewards.

Reward shaping and multi-component rewards: Beyond binary correctness, reward shaping techniques are often used in RL to guide agents via additional terms. In LLM training, some works combine fluency and factuality with task-specific rewards. For example, one might include a KL or perplexity penalty to keep outputs fluent. \textbf{Our continuous reward is in spirit a form of shaping: it rewards not only correct final answers but also low-perplexity (high-confidence) generations, explicit logical steps, and consistency between reasoning and answer.} Similar ideas appear in RL for LLMs and self-training methods. The idea of enforcing consistency in reasoning models (for example, through self-consistency) has also been extended into preference-based training settings where the preferences are derived automatically, without requiring human-labeled supervision. Unlike prior work, we combine multiple handcrafted metrics into one continuous reward function, and systematically compare it to simpler reward signals \cite{feng2025efficientreasoningmodelssurvey}.

\section{Methodology}

We fine-tune a pretrained causal LM using reinforcement learning with different reward schemes. We first describe the model and training algorithm, then detail each reward type and the hybrid scheduler.

\subsection{Model and Training Setup}

Our policy is the Qwen3-4B model (a 4B-parameter open LLM) with LoRA adapters activated. We tokenize prompts and ensure outputs contain an XML structure: each model output includes \texttt{<reasoning>...</reasoning><answer>...</answer>}. The \texttt{<answer>} tag should contain the final numeric answer. We prepare the GSM8K dataset of math problems, converting each to a prompt.

Training is done with a PPO-like optimizer. We use the Hugging Face TRL library’s \texttt{GRPOTrainer} (Group Relative PPO) which handles multiple sampled completions per prompt and normalizes rewards. At each training step we sample a batch of prompts. For each prompt we generate $G$ candidate outputs under the current policy. For each completion we compute a scalar reward (as described below). We then compute advantages by centering and scaling the rewards within each group:

\begin{equation}
A_i = \frac{r_i - \bar{r}}{\sigma}, 
\end{equation}

where $\bar{r}$ and $\sigma$ are the group mean and standard deviation. The policy is updated to maximize the RL objective with a KL-constraint to the reference (initial) policy. Hyperparameters (learning rate, epochs, etc.) are set as in GRPO reference implementations.

The training loop is summarized as:

\begin{algorithm}[H]
\caption{Training loop}
\begin{algorithmic}[1]
\For{$step = 1$ to $T$}
    \State Sample batch of prompts $\{p_j\}$
    \State Generate $G$ completions per prompt: $\{o_{j,g}\}$
    \For{each completion $o$}
        \State Extract final\_answer from \texttt{<answer>}
        \State Compute reward $R(o)$ according to chosen scheme
    \EndFor
    \State Normalize rewards $\to$ advantages (GRPO scaling)
    \State Perform PPO update on model using advantages
    \If{hybrid schedule}
        \State Update reward weights using scheduler
    \EndIf
    \State Log metrics (accuracy, avg reward, components)
\EndFor
\end{algorithmic}
\end{algorithm}

\subsection{Hard (Binary) Reward}

The hard reward assigns 1.0 if the model’s final answer exactly matches the ground truth, and 0 otherwise. Formally, let $\hat{y}$ be the predicted answer and $y^*$ the true answer:

\begin{equation}
R_{\text{correct}}(\hat{y}, y^*) =
\begin{cases}
1.0, & \hat{y} = y^* \\
0.0, & \text{otherwise}.
\end{cases}
\end{equation}

We also add a small format reward to encourage the XML structure:

\begin{equation}
R_{\text{format}}(o) =
\begin{cases}
v_f, & \text{if $o$ contains both <reasoning> and <answer> tags} \\
0.0, & \text{otherwise},
\end{cases}
\end{equation}

where $v_f$ is a fixed value (e.g.\ 0.2). The hard reward is then

\begin{equation}
R_{\text{hard}} = R_{\text{correct}} + R_{\text{format}}.
\end{equation}

This lies in $[0, 1+v_f]$, clamped to $\leq 1$.  

Example: If the reasoning/answer format is correct but the answer is wrong, $R_{\text{correct}}=0$ and $R_{\text{format}}=v_f>0$, yielding a partial reward.

\subsection{Continuous (Multi-component) Reward}

We design a continuous reward that combines four components: correctness, perplexity, reasoning quality, and consistency. Let $\omega_C, \omega_P, \omega_R, \omega_I$ be nonnegative weights (summing to $\leq 1$). Then

\begin{equation}
R_{\text{cont}} = \omega_C R_{\text{correct}}^{(cont)} + \omega_P R_{\text{perp}} + \omega_R R_{\text{reason}} + \omega_I R_{\text{consist}}.
\end{equation}

\paragraph{Correctness component.}  
For numeric answers, we compute relative error:
\begin{equation}
\epsilon = \frac{|\hat{y} - y^*|}{\max(|y^*|, 1)}.
\end{equation}

We also include order-of-magnitude similarity:
\begin{equation}
R_{\text{correct}}^{(cont)} = \alpha e^{-\epsilon} + \beta \frac{1}{1 + \left| \log_{10}|\hat{y}| - \log_{10}|y^*| \right|} + \gamma \cdot \mathbf{1}\left(|\log_{10}|\hat{y}| - \log_{10}|y^*|| < 1\right).
\end{equation}

If non-numeric, we use:
\begin{equation}
R_{\text{correct}}^{(cont)} = \alpha' \, SeqSim(\hat{y}, y^*) + \beta' \, WordOverlap(\hat{y}, y^*).
\end{equation}

\paragraph{Perplexity component.}  
Given normalized log-losses $\ell_{\text{full}}, \ell_{\text{reason}}, \ell_{\text{ans}}$:
\begin{equation}
R_{\text{perp}} = w_f e^{-\ell_{\text{full}}/\tau_f} + w_r e^{-\ell_{\text{reason}}/\tau_r} + w_a e^{-\ell_{\text{ans}}/\tau_a},
\end{equation}
with $w_f+w_r+w_a=1$.

\paragraph{Reasoning quality.}  
Based on length ($L$), step indicators ($S$), and math symbols ($M$):
\begin{equation}
R_{\text{reason}} = w_L L + w_S S + w_M M, \quad w_L+w_S+w_M=1.
\end{equation}

\paragraph{Consistency.}  
Agreement between reasoning and final answer:
\begin{equation}
R_{\text{consist}} =
\begin{cases}
C_{\text{match}}, & |\hat{y}-y^*|<\delta \\
C_{\text{num}}, & \text{both numeric but differ} \\
C_{\text{partial}}, & \text{one numeric only} \\
0, & d<0.2 \\
0.5, & 0.2 \leq d < 0.8 \\
1.0, & d \geq 0.8,
\end{cases}
\end{equation}
where $d$ is sequence similarity.

\subsection{Hybrid Reward and Scheduler}

We combine the hard and continuous rewards with time-varying weights. Let $w_{\text{hard}}(t), w_{\text{cont}}(t)$ be nonnegative weights (sum to 1) at training step $t$:

\begin{equation}
R_{\text{hybrid}}(t) = w_{\text{hard}}(t) R_{\text{hard}} + w_{\text{cont}}(t) R_{\text{cont}}.
\end{equation}

\paragraph{Continuous-to-hard schedule.}  
For $t<T_s$: $(w_{\text{cont}}, w_{\text{hard}})=(1,0)$.  
For $T_s \leq t < T_e$:
\begin{equation}
w_{\text{hard}}(t) = \frac{t - T_s}{T_e - T_s}, \quad w_{\text{cont}}(t) = 1 - w_{\text{hard}}(t).
\end{equation}
For $t \geq T_e$: $(w_{\text{cont}}, w_{\text{hard}})=(0,1)$.

\paragraph{Hard-to-continuous schedule.}  
The reverse of the above: $(0,1) \to (1,0)$.

\begin{algorithm}[H]
\caption{Hybrid Reward Training}
\begin{algorithmic}[1]
\State Initialize scheduler with $(T_s, T_e)$
\For{$step = 1$ to $T$}
    \State Compute $R_{\text{hard}}$ and $R_{\text{cont}}$
    \State Get weights from scheduler: $\{w_{\text{hard}}, w_{\text{cont}}\}$
    \State Combine: $R_{\text{hybrid}} = w_{\text{hard}} R_{\text{hard}} + w_{\text{cont}} R_{\text{cont}}$
    \State Clip $R_{\text{hybrid}} \in [0,1]$
    \State Update model via PPO/GRPO
\EndFor
\end{algorithmic}
\end{algorithm}

\subsection{Training Loop and Logging}

At each step, we log average reward, accuracy, perplexity, and each component. Stability is measured via variance of accumulated rewards:

\begin{equation}
\text{Stability} = \frac{1}{1 + \mathrm{Var}(R)}.
\end{equation}

A lower variance implies more stable optimization.

\section{Experiments}

We evaluate on the GSM8K dataset~\cite{cobbe2021trainingverifierssolvemath}, which contains 8,500 grade-school math problems with numeric answers. We split out a held-out test set of 100 problems for final evaluation. All experiments fine-tune Qwen3-4B (\textit{LoRA, base model from Unsloth}) for 200 training steps on 1000 training samples with batch size $1$ and $4$ completions per prompt. We use GRPO \cite{shao2024deepseekmathpushinglimitsmathematical} with a fixed learning rate of 
\begin{equation}
\eta = 5 \times 10^{-6}.
\end{equation}
The hybrid schedules use
\begin{equation}
T_{\text{start}} = 50, \quad T_{\text{end}} = 150.
\end{equation}
Random seed is fixed ($3407$) for reproducibility.

We compare four reward configurations:
\begin{itemize}
    \item \textbf{Hard:} Binary correctness + format (as in Sec.~3.2).
    \item \textbf{Continuous:} Multi-component reward (Sec.~3.3) with fixed component weights.
    \item \textbf{Hybrid (cont$\rightarrow$hard):} Continuous-to-hard schedule.
    \item \textbf{Hybrid (hard$\rightarrow$cont):} Hard-to-continuous schedule.
\end{itemize}

We report accuracy (fraction of exact correct answers) and training stability at the end of training. We also analyze intermediate training dynamics (accuracy and reward vs.\ step).

All experiments used identical initial policy (SFT fine-tuned for format) and hyperparameters; only the reward function differs. We evaluated final model accuracy by greedy decoding on 100 held-out GSM8K problems.

\section{Results}
\begin{figure}[htbp]
    \centering
    \includegraphics[width=1\linewidth]{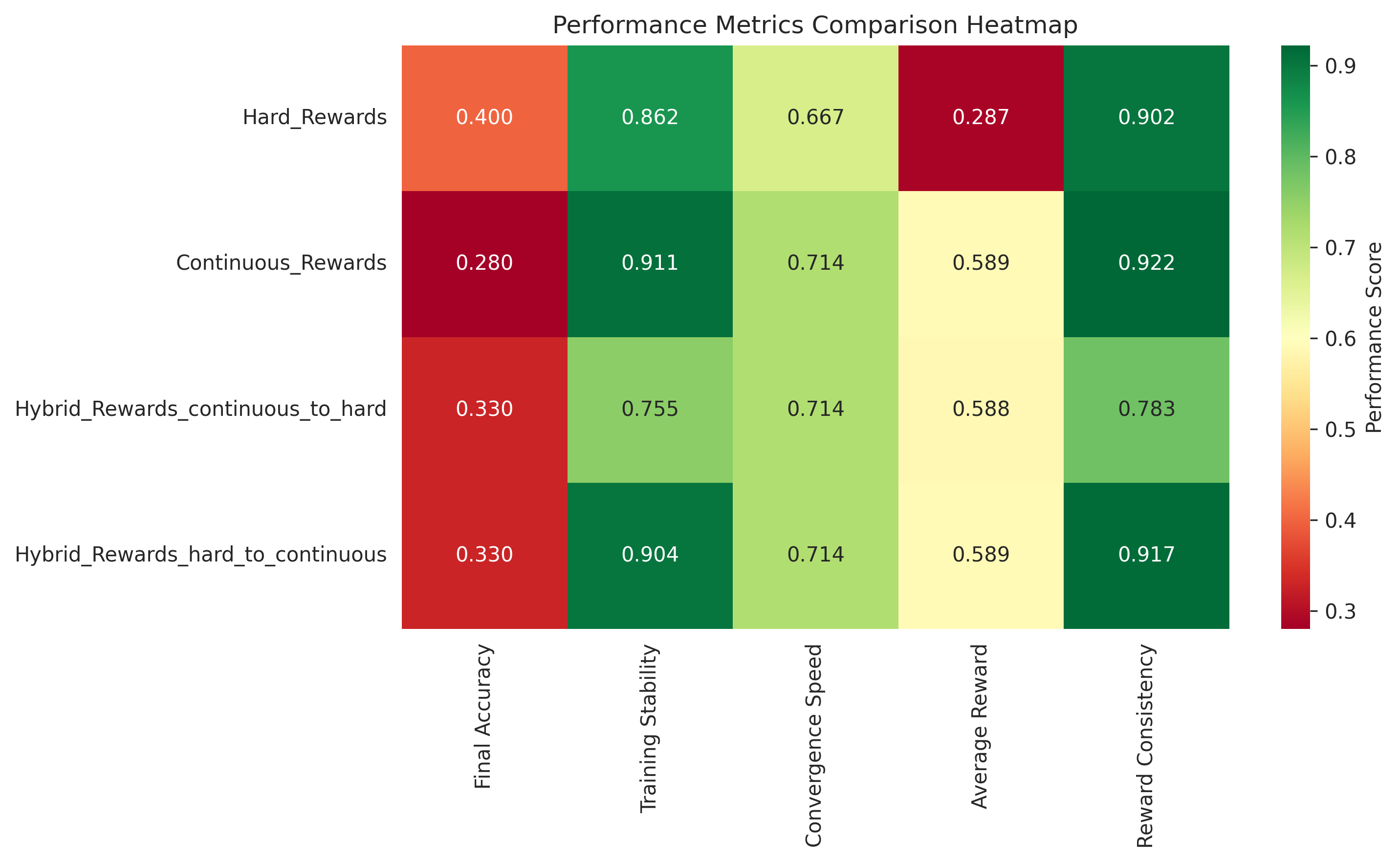}
    \caption{Performance Heatmap}
    \label{fig:Fig1}
\end{figure}

Table 1 summarizes the main results. The hard reward strategy achieved the highest final accuracy ($40\%$) and fastest convergence. Continuous reward lagged substantially ($28\%$). Both hybrid schedules were intermediate ($33\%$ accuracy). While the hard reward produced the highest final accuracy, it did not yield the lowest reward variance. Instead, the continuous reward exhibited the highest measured stability ($0.911$), suggesting a decoupling between optimization stability and task alignment: continuous shaping gives a smoother (lower-variance) learning signal that may encourage stable updates but can also dilute the objective and lead the model to optimize proxies (e.g., perplexity) rather than true correctness.

\begin{table}[h!]
\centering
\begin{tabular}{lccccc}
\toprule
\textbf{Method} & 
\makecell{\textbf{Final} \\ \textbf{Accuracy}} & 
\makecell{\textbf{Final} \\ \textbf{Perplexity}} & 
\makecell{\textbf{Conv.} \\ \textbf{Step}} & 
\makecell{\textbf{Training} \\ \textbf{Stability}} & 
\makecell{\textbf{Avg} \\ \textbf{Reward}} \\
\midrule
Hard\_Rewards & 0.400 & 2.18 & 5 & 0.862 & 0.287 \\
Continuous\_Rewards & 0.280 & 2.28 & 4 & 0.911 & 0.589 \\
\makecell{Hybrid \\ cont.$\to$hard} & 0.330 & 2.25 & 4 & 0.755 & 0.588 \\
\makecell{Hybrid \\ hard$\to$cont.} & 0.330 & 2.19 & 4 & 0.904 & 0.589 \\
\bottomrule
\end{tabular}
\caption{Performance metrics across different reward structures.}
\label{tab:reward_metrics}
\end{table}

Convergence step is defined as the training step at which maximum accuracy was first reached. The hard schedule converged by step 5, whereas continuous did converge by step 4. The scheduler balancing (cont$\rightarrow$hard vs.\ hard$\rightarrow$cont) gave similar final accuracy, but cont$\rightarrow$hard reached somehow highest Final Perplexity.

\begin{figure}[htbp]
    \centering
    \includegraphics[width=1\linewidth]{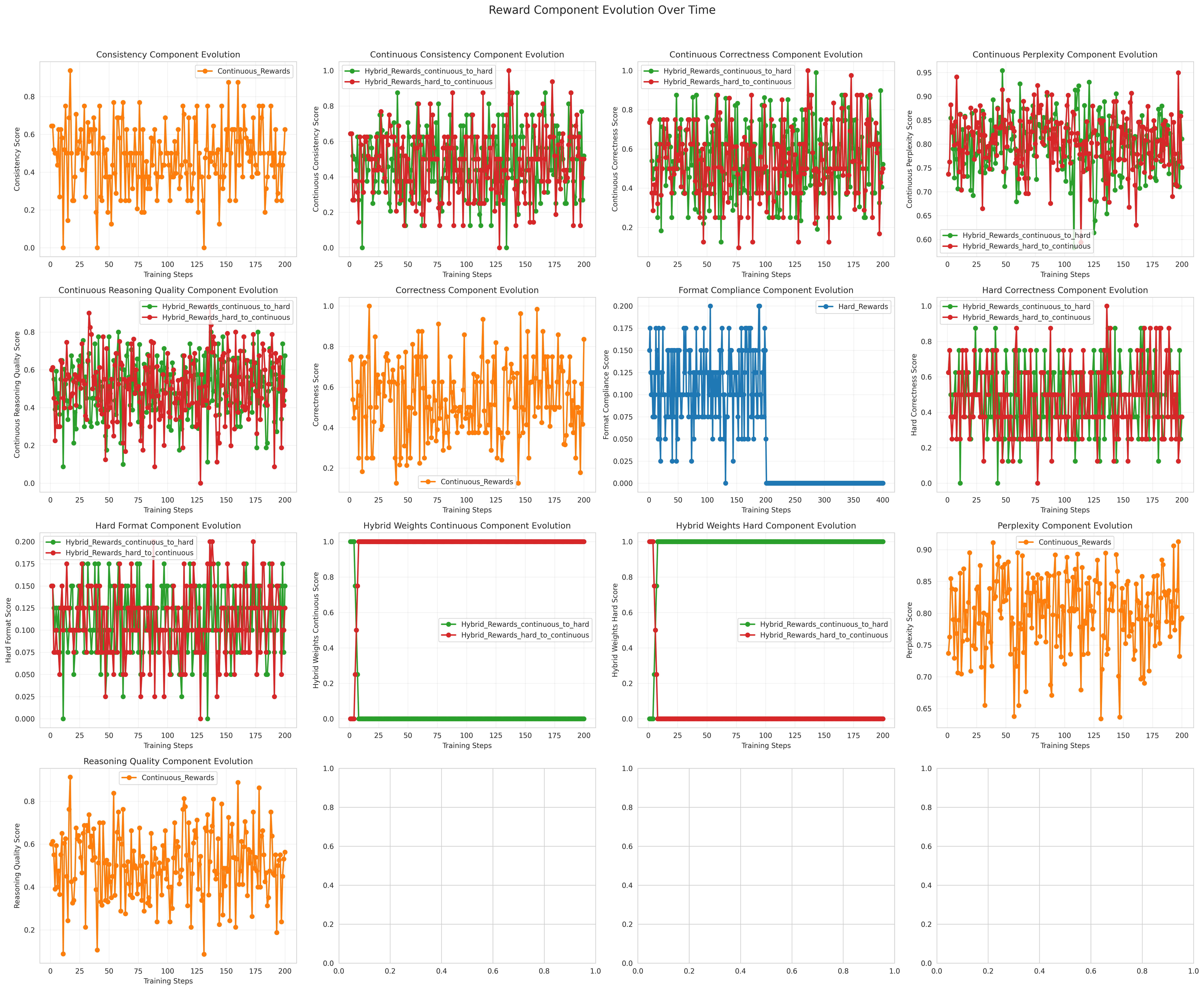}
    \caption{Reward Components Evolution}
    \label{fig:Fig2}
\end{figure}

\begin{figure}[htbp]
    \centering
    \includegraphics[width=1\linewidth]{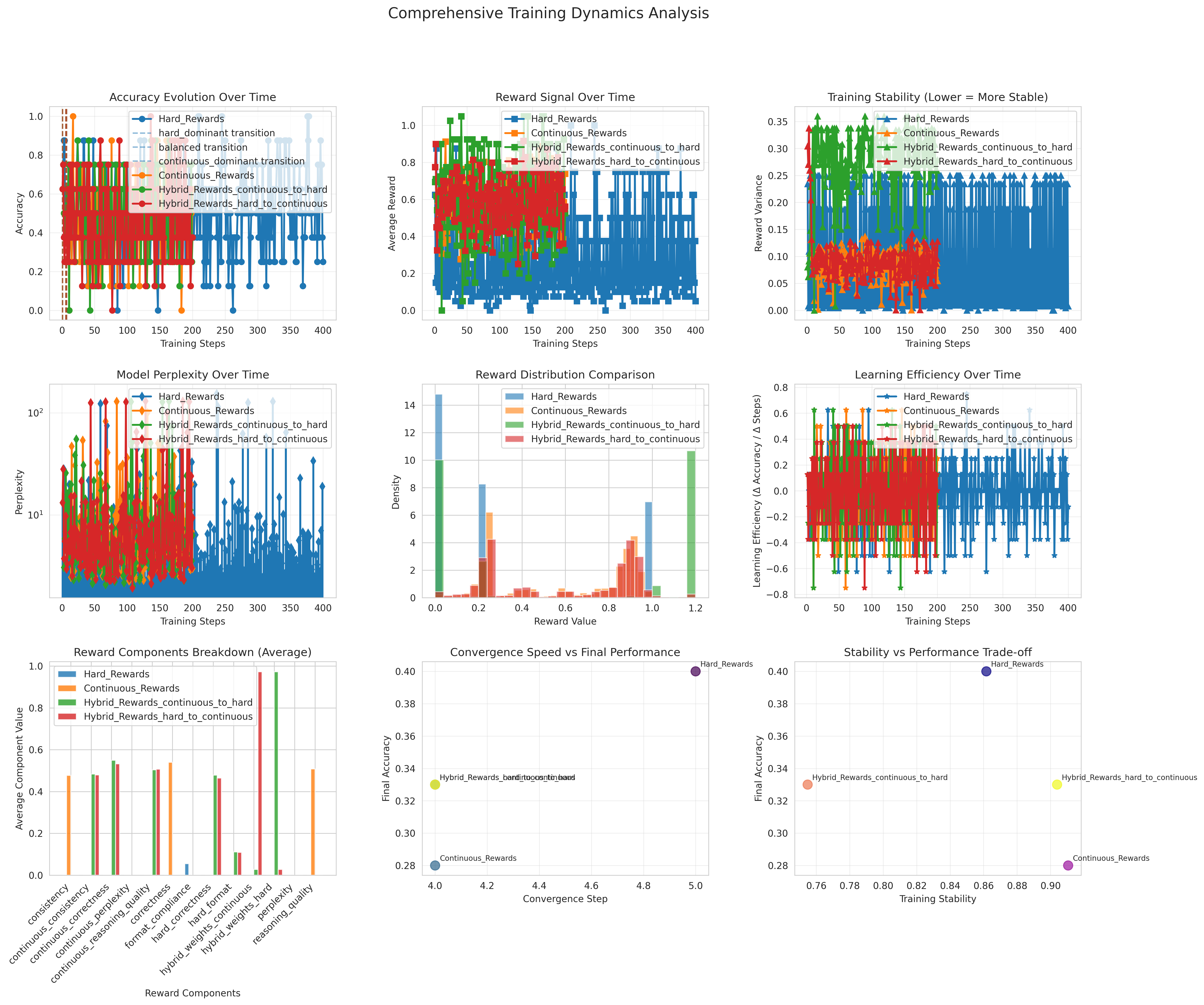}
    \caption{Training Dynamics Comprehensive}
    \label{fig:Fig3}
\end{figure}

Figure~\ref{fig:Fig3} (left) shows accuracy vs.\ training steps. The hard reward quickly rises to $\sim 0.4$ within 50 steps, then plateaus. Continuous reward rises slowly and unevenly. Hybrid methods start closer to continuous (since either begins with cont or hard) and eventually converge to $\sim 0.33$ as the weights shift. Figure~\ref{fig:Fig2} plots the evolution of reward components (e.g., correctness vs.\ perplexity) for the continuous experiment, showing that the model quickly maximized easier components (like perplexity) but struggled to improve true correctness.

\section{Discussion}

The empirical findings highlight several insights about reward design in LLM fine-tuning:

\textbf{Binary vs.\ Shaped Rewards:} The hard reward’s superior performance suggests that, for precise tasks like arithmetic, a direct accuracy signal is hard to beat. The continuous reward, despite richer information, may have provided conflicting objectives (e.g., encouraging fluency or verbosity) that diluted the focus on correctness. This aligns with the notion that RL agents can exploit unintended correlations in complex rewards ~\cite{ziegler2020finetuninglanguagemodelshuman}. In our case, the model may have optimized to minimize perplexity or produce ``\textit{reasonable-looking}'' reasoning chains without actually improving final answer accuracy.

\textbf{Training Stability:} The experimental results demonstrate that the continuous reward scheme exhibited the most stable reward value distribution, as reflected by its highest stability metric ($0.911$). In comparison, the hard reward scheme achieved slightly lower stability ($0.862$), indicating higher reward variance despite faster convergence and higher final accuracy. This suggests that while continuous rewards provided smoother and more consistent optimization dynamics, their multi-component design may have introduced competing optimization signals that hindered overall performance gains. The hybrid reward schedules offered an intermediate balance, partially mitigating variance while maintaining reasonable convergence behavior.

\textbf{Consistency and Alignment:} Our consistency component was motivated by the observation that more consistent answers are often correct~\cite{prasad2025selfconsistencypreferenceoptimization}. However, implementing this automatically is challenging. If the model rarely assigns multiple completions per prompt during training (GRPO uses a small group size), the notion of ``\textit{consistency across samples}'' is weak. We instead compared the reasoning’s final number to the answer’s number in a single output, which is a form of self-consistency check. While theoretically sensible, in practice this did not dramatically boost performance. 

\textbf{Reward Modeling Challenges:} Our results echo broader challenges in reward modeling: reward models for reasoning tasks often generalize poorly~\cite{gao2022scalinglawsrewardmodel}, and care must be taken that reward scores truly capture the intended behavior. We observed that some continuous components (e.g., step count, word count) can be gamed (the model might insert words like ``\textit{first…second…}'' without genuine logic). Without human-aligned preference data, shaping rewards remains heuristic.

\textbf{Motivation for Hybrid/Adaptive Strategies:} Although the hard reward performed best here, hybrid approaches offer flexibility. In other tasks or later in training, it might be useful to allow the model to explore under a softer reward and then refine with a strict one (or vice versa). Adaptive schedules can also serve as a form of curriculum learning on the reward itself. Our formal scheduler provides a blueprint for such curricula.

\textbf{Extensibility:} Our framework is designed to add new reward terms easily. For creative tasks (e.g., essay writing, poetry), one could plug in an aesthetic or style reward and schedule it similarly. The logging infrastructure (per-step metrics for each component) helps diagnose which signals are influencing the model.

In summary, while our continuous reward structure was intuitively appealing, the experiment underscores that alignment to the true goal (correct answer) is paramount. Future work could explore learned reward models (e.g., trained classifiers on reasoning correctness) instead of hand-crafted rules, incorporate human feedback, or use more sophisticated multi-objective methods~\cite{lambert2024rewardbenchevaluatingrewardmodels}.

\section{Limitations}
While our framework highlights the importance of reward structure design, several limitations remain. First, our experiments were conducted at a relatively small scale, focusing on Qwen3-4B with LoRA fine-tuning on GSM8K. The extent to which our findings transfer to larger foundation models or to broader domains, particularly those involving continuous or hybrid reward signals, is still uncertain. Second, the continuous reward functions employed in our setup rely on handcrafted metrics such as perplexity and reasoning length, which serve only as imperfect proxies for reasoning quality. Any misalignment in these metrics can introduce reward misspecification, potentially steering the model toward undesirable behaviors. Third, our training horizon was short, capped at a few hundred optimization steps for feasibility. Longer training schedules may surface stability issues or reveal alternative convergence patterns not captured in our current results. In terms of evaluation, we primarily relied on exact-match accuracy and reward variance; human-centric evaluations of reasoning clarity, interpretability, and faithfulness were not performed, limiting the scope of our assessment. Finally, the hybrid reward scheduler was predefined and fixed in its design. More flexible or adaptive mechanisms could yield better performance, and our rigid linear transition likely underutilizes the potential of hybrid reward shaping.

\section{Future Work}
Our findings open up several promising avenues for further research. Scaling the experiments to larger foundation models with tens of billions of parameters, as well as applying the framework across more diverse reasoning datasets, would provide stronger evidence of generality. Another natural extension is to move beyond hand-designed hybrid reward schedules and explore reinforcement meta-learning approaches that dynamically adapt the weighting of discrete and continuous components during training. Human-in-the-loop alignment also represents an important direction: integrating expert judgments or preference modeling into the continuous reward channel could reduce dependence on noisy, handcrafted proxies. Beyond mathematics, the hybrid reward paradigm could be extended to creative domains such as dialogue, poetry, or program synthesis, where the balance between correctness and quality introduces new challenges.

\section{Conclusion}

We have introduced and rigorously evaluated adaptive reward schemes for RL fine-tuning of language models on mathematical reasoning. Our contributions include detailed mathematical formulations of hard, continuous, and hybrid rewards (with formal scheduler equations), as well as an empirical comparison on GSM8K using a LoRA-adapted Qwen3 model. The main takeaway is that reward design critically affects alignment and performance: a simple binary correctness reward outperformed our more complex shaping approach in this setting. This highlights the alignment challenge: more reward signals do not automatically yield better aligned behavior without careful calibration. We provided pseudocode and logging strategies to ensure all findings are reproducible. In the future, integrating learned reward models, human preferences, or more elaborate curricula could further improve reasoning alignment.

\section*{Societal Impact and Scalable Oversight Integration}

This work advances scalable oversight by showing how hybrid reward signals can align increasingly capable LLMs without heavy human supervision. The approach blends binary correctness with process-based signals, allowing models to be guided even when ground-truth labels are unavailable. This naturally supports oversight methods such as \textit{debate, recursive reward modeling, and constitutional AI}, where intermediate reasoning must be evaluated. The hybrid scheduler provides a principled way to shift from sparse verification to denser shaping rewards, reducing labeling costs while maintaining alignment pressure. We note dual-use risks: automated reward shaping can increase reward hacking if not carefully monitored. Integrating this framework with iterated amplification or AI-assisted evaluation may support safe deployment in high-stakes domains.

\section*{Acknowledgements}
We would like to thank Philip Quirke, Amir Abdullah, and Jacob Haimes for their valuable discussions, feedback, and support throughout this work.

\bibliography{references}

@misc{lambert2024rewardbenchevaluatingrewardmodels,
      title={RewardBench: Evaluating Reward Models for Language Modeling}, 
      author={Nathan Lambert and Valentina Pyatkin and Jacob Morrison and LJ Miranda and Bill Yuchen Lin and Khyathi Chandu and Nouha Dziri and Sachin Kumar and Tom Zick and Yejin Choi and Noah A. Smith and Hannaneh Hajishirzi},
      year={2024},
      eprint={2403.13787},
      archivePrefix={arXiv},
      primaryClass={cs.LG},
      url={https://arxiv.org/abs/2403.13787}, 
}

@misc{prasad2025selfconsistencypreferenceoptimization,
      title={Self-Consistency Preference Optimization}, 
      author={Archiki Prasad and Weizhe Yuan and Richard Yuanzhe Pang and Jing Xu and Maryam Fazel-Zarandi and Mohit Bansal and Sainbayar Sukhbaatar and Jason Weston and Jane Yu},
      year={2025},
      eprint={2411.04109},
      archivePrefix={arXiv},
      primaryClass={cs.CL},
      url={https://arxiv.org/abs/2411.04109}, 
}

@misc{ziegler2020finetuninglanguagemodelshuman,
      title={Fine-Tuning Language Models from Human Preferences}, 
      author={Daniel M. Ziegler and Nisan Stiennon and Jeffrey Wu and Tom B. Brown and Alec Radford and Dario Amodei and Paul Christiano and Geoffrey Irving},
      year={2020},
      eprint={1909.08593},
      archivePrefix={arXiv},
      primaryClass={cs.CL},
      url={https://arxiv.org/abs/1909.08593}, 
}

@misc{cobbe2021trainingverifierssolvemath,
      title={Training Verifiers to Solve Math Word Problems}, 
      author={Karl Cobbe and Vineet Kosaraju and Mohammad Bavarian and Mark Chen and Heewoo Jun and Lukasz Kaiser and Matthias Plappert and Jerry Tworek and Jacob Hilton and Reiichiro Nakano and Christopher Hesse and John Schulman},
      year={2021},
      eprint={2110.14168},
      archivePrefix={arXiv},
      primaryClass={cs.LG},
      url={https://arxiv.org/abs/2110.14168}, 
}

@misc{shao2024deepseekmathpushinglimitsmathematical,
      title={DeepSeekMath: Pushing the Limits of Mathematical Reasoning in Open Language Models}, 
      author={Zhihong Shao and Peiyi Wang and Qihao Zhu and Runxin Xu and Junxiao Song and Xiao Bi and Haowei Zhang and Mingchuan Zhang and Y. K. Li and Y. Wu and Daya Guo},
      year={2024},
      eprint={2402.03300},
      archivePrefix={arXiv},
      primaryClass={cs.CL},
      url={https://arxiv.org/abs/2402.03300}, 
}

@misc{yang2025qwen3technicalreport,
      title={Qwen3 Technical Report}, 
      author={An Yang and Anfeng Li and Baosong Yang and Beichen Zhang and Binyuan Hui and Bo Zheng and Bowen Yu and Chang Gao and Chengen Huang and Chenxu Lv and Chujie Zheng and Dayiheng Liu and Fan Zhou and Fei Huang and Feng Hu and Hao Ge and Haoran Wei and Huan Lin and Jialong Tang and Jian Yang and Jianhong Tu and Jianwei Zhang and Jianxin Yang and Jiaxi Yang and Jing Zhou and Jingren Zhou and Junyang Lin and Kai Dang and Keqin Bao and Kexin Yang and Le Yu and Lianghao Deng and Mei Li and Mingfeng Xue and Mingze Li and Pei Zhang and Peng Wang and Qin Zhu and Rui Men and Ruize Gao and Shixuan Liu and Shuang Luo and Tianhao Li and Tianyi Tang and Wenbiao Yin and Xingzhang Ren and Xinyu Wang and Xinyu Zhang and Xuancheng Ren and Yang Fan and Yang Su and Yichang Zhang and Yinger Zhang and Yu Wan and Yuqiong Liu and Zekun Wang and Zeyu Cui and Zhenru Zhang and Zhipeng Zhou and Zihan Qiu},
      year={2025},
      eprint={2505.09388},
      archivePrefix={arXiv},
      primaryClass={cs.CL},
      url={https://arxiv.org/abs/2505.09388}, 
}

@misc{ouyang2022traininglanguagemodelsfollow,
      title={Training language models to follow instructions with human feedback}, 
      author={Long Ouyang and Jeff Wu and Xu Jiang and Diogo Almeida and Carroll L. Wainwright and Pamela Mishkin and Chong Zhang and Sandhini Agarwal and Katarina Slama and Alex Ray and John Schulman and Jacob Hilton and Fraser Kelton and Luke Miller and Maddie Simens and Amanda Askell and Peter Welinder and Paul Christiano and Jan Leike and Ryan Lowe},
      year={2022},
      eprint={2203.02155},
      archivePrefix={arXiv},
      primaryClass={cs.CL},
      url={https://arxiv.org/abs/2203.02155}, 
}

@misc{hu2021loralowrankadaptationlarge,
      title={LoRA: Low-Rank Adaptation of Large Language Models}, 
      author={Edward J. Hu and Yelong Shen and Phillip Wallis and Zeyuan Allen-Zhu and Yuanzhi Li and Shean Wang and Lu Wang and Weizhu Chen},
      year={2021},
      eprint={2106.09685},
      archivePrefix={arXiv},
      primaryClass={cs.CL},
      url={https://arxiv.org/abs/2106.09685}, 
}

@misc{wei2023chainofthoughtpromptingelicitsreasoning,
      title={Chain-of-Thought Prompting Elicits Reasoning in Large Language Models}, 
      author={Jason Wei and Xuezhi Wang and Dale Schuurmans and Maarten Bosma and Brian Ichter and Fei Xia and Ed Chi and Quoc Le and Denny Zhou},
      year={2023},
      eprint={2201.11903},
      archivePrefix={arXiv},
      primaryClass={cs.CL},
      url={https://arxiv.org/abs/2201.11903}, 
}

@misc{feng2025efficientreasoningmodelssurvey,
      title={Efficient Reasoning Models: A Survey}, 
      author={Sicheng Feng and Gongfan Fang and Xinyin Ma and Xinchao Wang},
      year={2025},
      eprint={2504.10903},
      archivePrefix={arXiv},
      primaryClass={cs.CL},
      url={https://arxiv.org/abs/2504.10903}, 
}

@misc{gao2022scalinglawsrewardmodel,
      title={Scaling Laws for Reward Model Overoptimization}, 
      author={Leo Gao and John Schulman and Jacob Hilton},
      year={2022},
      eprint={2210.10760},
      archivePrefix={arXiv},
      primaryClass={cs.LG},
      url={https://arxiv.org/abs/2210.10760}, 
}
\bibliographystyle{unsrtnat}

\appendix

\section{Comprehensive RL Reward Structure Analysis}

\subsection{Analysis by Reward Type}

\paragraph{Hard Rewards}
\begin{itemize}
    \item Best: \textbf{Hard\_Rewards}, accuracy = 0.400, perplexity = 2.18.
    \item Converged at step 5.
    \item Binary signal $\Rightarrow$ fast convergence but limited nuance.
\end{itemize}

\paragraph{Continuous Rewards}
\begin{itemize}
    \item Best: \textbf{Continuous\_Rewards}, accuracy = 0.280, perplexity = 2.28.
    \item Converged at step 4.
    \item Smooth signal $\Rightarrow$ stable learning and partial credit.
\end{itemize}

\paragraph{Hybrid Rewards}
\begin{itemize}
    \item Best: \textbf{Hybrid\_cont.\,$\to$\,hard}, accuracy = 0.330, perplexity = 2.25.
    \item Converged at step 4.
    \item Adaptive $\Rightarrow$ combines hard/continuous benefits.
\end{itemize}

\subsection{Key Findings}
\begin{enumerate}
    \item \textbf{Highest Accuracy:} Hard\_Rewards (0.400).
    \item \textbf{Most Stable Training:} Continuous\_Rewards (0.911).
    \item \textbf{Fastest Convergence:} Continuous\_Rewards (step 4).
\end{enumerate}

\subsection{Statistical Analysis}

The T-test and Cohen’s $d$ were computed as:
\begin{equation}
t = \frac{\bar{x}_1 - \bar{x}_2}{s_p \sqrt{2/n}}, 
\quad 
d = \frac{\bar{x}_1 - \bar{x}_2}{s_p},
\end{equation}
where $s_p$ is the pooled standard deviation.

\begin{itemize}
    \item Hard vs. Continuous: $t=-28.63$, $p=0.0000$, $d=-0.809$ (Large).
    \item Hard vs. Hybrid (cont.\,$\to$\,hard): $t=-18.93$, $p=0.0000$, $d=-0.649$ (Medium).
    \item Hard vs. Hybrid (hard\,$\to$\,cont.): $t=-27.96$, $p=0.0000$, $d=-0.800$ (Large).
    \item Continuous vs. Hybrid (cont.\,$\to$\,hard): $t=0.077$, $p=0.939$, $d=0.003$ (Negligible).
    \item Continuous vs. Hybrid (hard\,$\to$\,cont.): $t=0.042$, $p=0.966$, $d=0.001$ (Negligible).
    \item Hybrid (cont.\,$\to$\,hard) vs. Hybrid (hard\,$\to$\,cont.): $t=-0.047$, $p=0.963$, $d=-0.002$ (Negligible).
\end{itemize}

\subsection{Recommendations}

\textbf{Hard\_Rewards (Weighted Score: 0.618)}  
\begin{itemize}
    \item Simple, binary, effective.  
    \item Clear optimization target $\Rightarrow$ faster learning in well-defined problems.  
    \item Limitation: weak handling of partial correctness.  
\end{itemize}

\paragraph{Use Cases}
\begin{itemize}
    \item \textbf{Hard Rewards:} Binary correctness, simple domains, fast convergence.  
    \item \textbf{Continuous Rewards:} Partial credit, stability, complex reasoning.  
    \item \textbf{Hybrid Rewards:} Adaptive tasks, exploration/exploitation balance.  
\end{itemize}

\section{Key Experimental Insights}

\paragraph{Performance Winners}
\begin{itemize}
    \item Highest Accuracy: Hard\_Rewards (0.400)
    \item Most Stable: Continuous\_Rewards (0.911)
    \item Fastest Convergence: Continuous\_Rewards (step 4)
\end{itemize}

\paragraph{Reward Type Analysis}
\begin{itemize}
    \item Hard Rewards: Accuracy = 0.400
    \item Continuous Rewards: Accuracy = 0.280
    \item Hybrid Rewards: Accuracy = 0.330
\end{itemize}

\paragraph{Training Dynamics}
\begin{itemize}
    \item Hard\_Rewards: Train $0.625 \to 0.250$, Eval = 0.400
    \item Continuous\_Rewards: Train $0.625 \to 0.625$, Eval = 0.280
    \item Hybrid (cont.\,$\to$\,hard): Train $0.625 \to 0.375$, Eval = 0.330
    \item Hybrid (hard\,$\to$\,cont.): Train $0.625 \to 0.375$, Eval = 0.330
\end{itemize}

\section{Experiment Configuration}

\lstdefinestyle{mystyle}{
  backgroundcolor=\color{gray!10},
  commentstyle=\color{green!40!black},
  keywordstyle=\color{blue!80!black},
  numberstyle=\tiny\color{gray},
  stringstyle=\color{orange!90!black},
  basicstyle=\ttfamily\footnotesize,
  breaklines=true,
  numbers=left,
  numbersep=5pt,
  frame=single,
  rulecolor=\color{black!40},
  tabsize=2,
  showstringspaces=false,
  captionpos=b
}

\lstset{style=mystyle}

\begin{lstlisting}[language=Python, caption={Centralized experiment configuration.}]
class ExperimentConfig:
    """Centralized configuration for the entire experiment."""
    # Model and LoRA
    MODEL_NAME: str = "unsloth/Qwen3-4B-Base"
    MAX_SEQ_LENGTH: int = 1024
    LORA_RANK: int = 8
    GPU_MEMORY_UTILIZATION: float = 0.5
    LORA_TARGET_MODULES: List[str] = field(default_factory=lambda: [
        "q_proj", "k_proj", "v_proj", "o_proj",
        "gate_proj", "up_proj", "down_proj",
    ])
    RANDOM_STATE: int = 3407

    # Dataset
    DATASET_TRAIN_LIMIT: int = 1000
    DATASET_EVAL_LIMIT: int = 100
    SYSTEM_PROMPT: str = """
Respond in the following format:
<reasoning>
Show your step-by-step mathematical reasoning here.
</reasoning>
<answer>
Provide the final numerical answer here.
</answer>
"""
    DOMAIN: str = "gsm8k"

    # Evaluation
    EVAL_BATCH_SIZE: int = 16
    MAX_NEW_TOKENS_EVAL: int = 200
    EVAL_TEMPERATURE: float = 0.7

    # Reward Function Parameters
    PERPLEXITY_CAP: float = 1000.0
    PERPLEXITY_MAX_LENGTH: int = 512
    PERPLEXITY_WEIGHTS: Dict[str, float] = field(default_factory=lambda: {
        'full_response': 0.4, 'reasoning': 0.3, 'answer': 0.3
    })
    PERPLEXITY_DECAY_FACTORS: Dict[str, float] = field(default_factory=lambda: {
        'full_response': 100.0, 'reasoning': 80.0, 'answer': 60.0
    })

    CORRECTNESS_WEIGHTS: Dict[str, float] = field(default_factory=lambda: {
        'relative_error': 0.6, 'magnitude_similarity': 0.25, 'order_of_magnitude': 0.15,
        'sequence_similarity': 0.7, 'word_overlap': 0.3
    })
    NUMERIC_CONSISTENCY_TOLERANCE: float = 0.01

    REASONING_LENGTH_MIN_WORDS: int = 20
    REASONING_LENGTH_MAX_WORDS: int = 200
    REASONING_LENGTH_IDEAL_WORDS: int = 100
    REASONING_LENGTH_PENALTY_FACTOR: float = 100.0
    REASONING_STEP_INDICATOR_THRESHOLD: int = 3
    REASONING_MATH_INDICATOR_THRESHOLD: int = 5
    REASONING_QUALITY_WEIGHTS: Dict[str, float] = field(default_factory=lambda: {
        'length_appropriateness': 0.4, 'step_indicators': 0.3, 'math_indicators': 0.3
    })

    CONSISTENCY_PARTIAL_REWARD: float = 0.3
    CONSISTENCY_MATCH_REWARD: float = 1.0
    CONSISTENCY_NUM_MATCH_REWARD: float = 0.5

    # Cell 4 continued: Configuration
    SOPHISTICATED_REWARD_COMPONENT_WEIGHTS: Dict[str, float] = field(default_factory=lambda: {
        'correctness': 0.4, 'perplexity': 0.25, 'reasoning_quality': 0.2, 'consistency': 0.15
    })

    # Hybrid Reward Schedule
    HYBRID_TRANSITION_STEPS: Tuple[int, int] = (3, 7)
    HARD_FORMAT_REWARD_VALUE: float = 0.2

    # GRPO Training Parameters
    GRPO_LEARNING_RATE: float = 5e-6
    GRPO_ADAM_BETA1: float = 0.9
    GRPO_ADAM_BETA2: float = 0.99
    GRPO_WEIGHT_DECAY: float = 0.1
    GRPO_WARMUP_RATIO: float = 0.1
    GRPO_LR_SCHEDULER_TYPE: str = "cosine"
    GRPO_OPTIM: str = "adamw_8bit"
    GRPO_LOGGING_STEPS: int = 1
    GRPO_PER_DEVICE_TRAIN_BATCH_SIZE: int = 1
    GRPO_GRADIENT_ACCUMULATION_STEPS: int = 4
    GRPO_NUM_GENERATIONS: int = 2
    GRPO_MAX_PROMPT_LENGTH: int = 256
    GRPO_MAX_COMPLETION_LENGTH: int = 300
    GRPO_MAX_STEPS: int = 200
    GRPO_SAVE_STEPS: int = 50
    GRPO_EVAL_STEPS: int = 50
    GRPO_MAX_GRAD_NORM: float = 0.1
    GRPO_REPORT_TO: str = "none"
    GRADIENT_CHECKPOINTING: bool = True
    USE_REENTRANT_CHECKPOINTING: bool = False

    # Visualization and Analysis
    HEATMAP_CONVERGENCE_NORMALIZATION_FACTOR: float = 10.0
    HEATMAP_REWARD_VARIANCE_NORMALIZATION_FACTOR: float = 1.0
    ANALYSIS_ACCURACY_IMPROVEMENT_THRESHOLD: float = 0.01
    ANALYSIS_ACCURACY_IMPROVEMENT_STEPS: int = 3
    ANALYSIS_P_VALUE_THRESHOLD: float = 0.05
    ANALYSIS_COHENS_D_SMALL: float = 0.2
    ANALYSIS_COHENS_D_MEDIUM: float = 0.5
    ANALYSIS_COHENS_D_LARGE: float = 0.8
    ANALYSIS_WEIGHTED_SCORE_WEIGHTS: Dict[str, float] = field(default_factory=lambda: {
        'accuracy': 0.4, 'stability': 0.3, 'convergence_speed': 0.3
    })

    # Creative Domain Rewards
    CREATIVE_POETRY_IDEAL_LENGTH: int = 50
    CREATIVE_JOKES_IDEAL_LENGTH: int = 30
    CREATIVE_REWARD_WEIGHTS: Dict[str, float] = field(default_factory=lambda: {
        'length': 0.25, 'structure': 0.25, 'creativity': 0.25, 'fluency': 0.25
    })
    CREATIVE_FLUENCY_DECAY_FACTOR: float = 1.1
\end{lstlisting}

\end{document}